\documentclass[leqno]{article}%
\usepackage{amsfonts,amsmath,amsthm,amssymb,times}
\usepackage{color}
\usepackage{graphicx,subfigure,color,float,enumerate}
\usepackage[utf8]{inputenc}
\usepackage{adjustbox}
\usepackage{mathptmx}
\usepackage[11pt]{moresize}
\pdfoutput=1
\newtheorem{definition}{DEFINITION}[section]

\newtheorem{axiom}{AXIOM} [section]

\theoremstyle{remark}

\numberwithin{equation}{section}

\begin{document}
\title{Nonparametric Data Analysis on the Space of Perceived Colors}
\author{Vic Patrangenaru and Yifang Deng\\
Florida State University, Tallahassee, FL, USA.}

\maketitle

\begin{abstract}
Moving around in a 3D world, requires the visual system of a living individual to rely on three channels of image recognition, which is done through three types of retinal cones. Newton, Grasmann, Helmholz and Schr$\ddot{o}$dinger laid down the basic assumptions needed to understand colored vision. Such concepts were furthered by Resnikoff, who imagined the space of perceived colors as a 3D homogeneous space.
 This article is concerned with perceived colors regarded as random objects on a Resnikoff 3D homogeneous space model. Two applications to color differentiation in machine vision are illustrated for the proposed statistical methodology, applied to the Euclidean model for perceived colors.
\end{abstract}


\section{Human vs Machine Color Perception}\label{s1}
Light, or the visible part of the electromagnetic radiation spectrum,
is the medium through which human beings receive the information  of surroundings. The physical nature of light is  electromagnetic radiation of different wavelengths ($\lambda$) and intensity($\varphi(\lambda)$). The human eye perceives light with different wavelengths as different colors, as long as the variation of wavelength is limited to the range between 370 $nm$ and 730 $nm$ (see Livingstone and Hubel(2008) \cite{MaDa:2008}).

From the perspective of physiology, the perception of color is formed in our brain by the superposition of the neural signals from three different kinds of photoreceptors which are distributed over the retina of human eye’s. The retina includes several layers of neural cells, beginning with the photoreceptors, the rods and cones. The main distinction between rods and cones is in their visual function. Rods serve vision at low luminance levels (e.g., less than $10^{-3}$ candela per square meter)
while cones serve vision at luminance levels higher than that. (see Fairchild(2005)\cite{Fa:2005}, Provenzi(2018)\cite{Pr:2018})In physiology, the receptors are called blue cones, red cones and green cones as they are stimulated by different wavelengths of electromagnetic radiation, thus transforming them to blue, red and green colors respectfully. Monochromatic light might also excite two types of cones simultaneously, thus producing the perception of another color. For example the electromagnetic radiation with wavelength of 580 $nm$ will produce a yellow color in our brain.

Another interesting fact is that our visual system cannot discriminate between monochromatic and broadband radiation. A spectral decomposition
of white light produces the perception of a mix of different-colored lights as experimentally proven by Newton\cite{Ne:1993}. In another words, the superposition of different lights generate a white color in our brain. Its like the group operator in abstract algebra.

Whereas color addition describes the perception of different colors caused by a superposition of red, green and blue light sources, the concept of color subtraction is based on the absorption or reflection  of objects with different surfaces. For example, an object with yellow color absorbed wavelength below around 500 $nm$.

According to the  physiology considerations above, every color which can be perceived by a healthy  human eye can be described by three numbers which measure the stimulation of red, green and blue cones. In the late twenties last century, Wright and Guild performed experiments on observers in color perception matched the color perception produced by monochromatic light (see Wright(1928) \cite{Wr:1928}, Guild(1932)\cite{Gu:1932}). Evaluation of these experiments resulted in the definition of the standardised RGB color matching functions, which have been transformed into the famous  CIE 1931 XYZ color matching functions, $\tilde{x}(\lambda)$, $\tilde{y}(\lambda)$, $\tilde{z}(\lambda)$. The XYZ tristimulus values of a certain spectral color is:
\begin{equation}
\begin{aligned}
X=k \int_{\lambda} \varphi(\lambda) \cdot  \tilde{x} (\lambda) d\lambda \\
Y=k \int_{\lambda} \varphi(\lambda) \cdot  \tilde{y} (\lambda) d\lambda \\
Z=k \int_{\lambda} \varphi(\lambda) \cdot  \tilde{z} (\lambda) d\lambda
\end{aligned}
\end{equation}
where $\varphi(\lambda)$  is the  spectral radiance.

\begin{figure}[h]
\centering
\includegraphics[scale=.5]{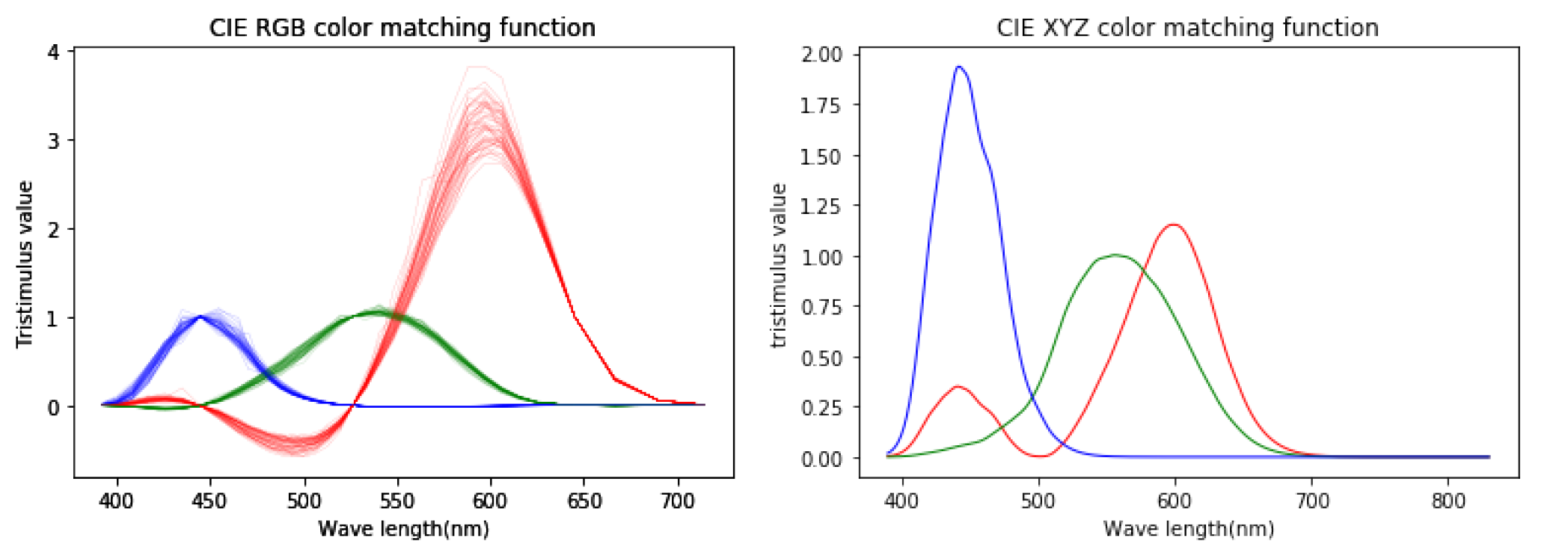}
\caption{Left side - RGB color matching function(CMF) based on 53 individuals perceptions. Right side - XYZ color matching function transformed from RGB CMF. 2 CMF are equivalent, and differed by a linear transformation.(Data from http://cvrl.ioo.ucl.ac.uk/)}
\label{RGB_XYZ}
\end{figure}

To represent color properly on a electronic device, such as on a monitor, printer or world wide web device, the standard RGB color space has been created by HP and Microsoft corporation in 1996 (see IEC Webstore \cite{We:1999}). The CIE RGB color space can be transformed to the sRGB color space by a linear transformation $M$ and a $\gamma$ correction, where
\begin{equation}
M=
\begin{bmatrix}
+3.2406&-1.5372&-0.4986 \\
-0.9689&+1.8758&+0.0415\\
+0.0557&-0.2040&+1.0570
\end{bmatrix}
\end{equation}

\begin{equation}
\gamma(u)=\left\{
\begin{aligned}
\frac{323u}{25},       u\leq 0.0031308 \\
\frac{211u^{\frac{5}{12}}-11}{200},  otherwise
\end{aligned}
\right.
\end{equation}
where $u$ is a value from the CIE XYZ color space after the transformation of $M$.

With the values of $R_{sRGB},G_{sRGB},B_{sRGB}$ , we can match the color properly in the $sRGB$ color space.

\begin{figure}[h]
\centering
\includegraphics[scale=0.2]{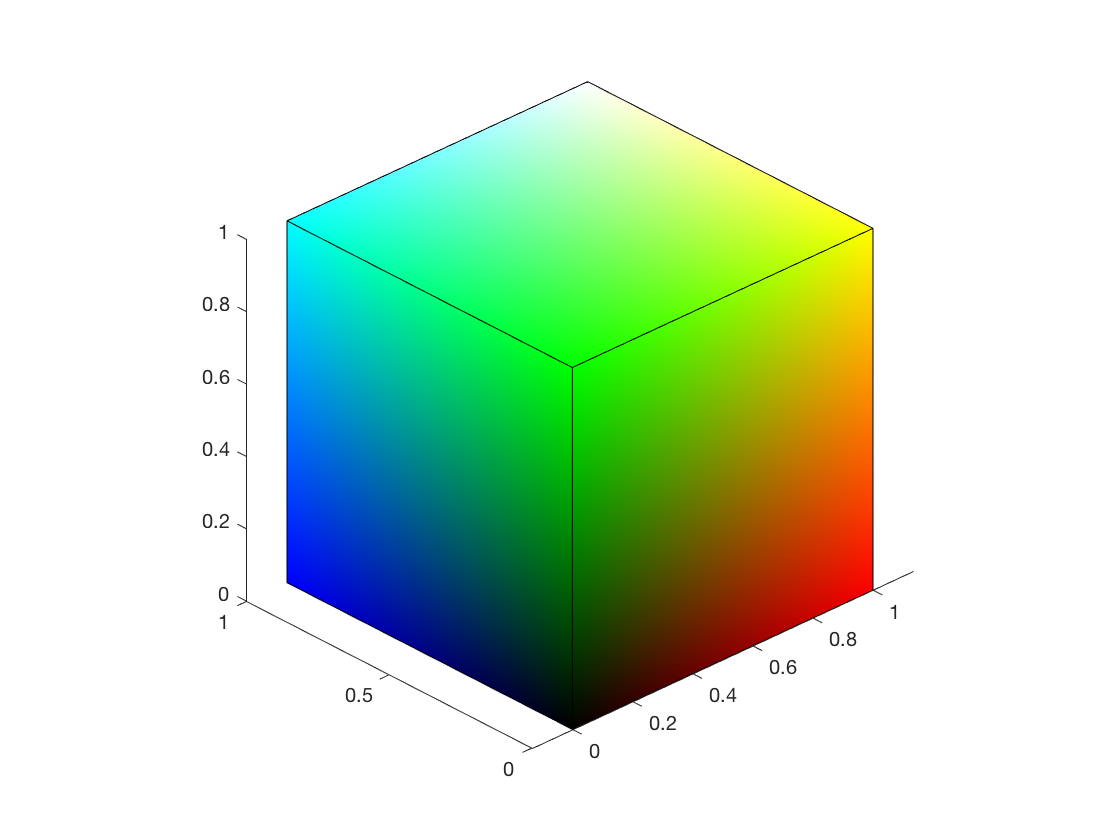}
\caption{XYZ color space}
\label{XYZCubic}
\end{figure}

\section{Motivation for Resnikoff's space of perceived colors}

 In the line of some of the considerations in Section \ref{s1}, Schr$\ddot{o}$dinger(1920)\cite{Sc:1920} proposed four axioms for the space of {\em perceived light}. A ray of light originating from a visible scene, also called {\em visible light} was imagined as a positive function $x:\Lambda =[a, b]\to \mathbb R_+,$ that is square integrable, that is $x \in L^2(\Lambda);$ here $a,b$ are the lower and upper bound of wavelengths of visible light.
 Let $S_i(\lambda), i=1,2,3$ denotes the spectral sensitivity associated with the visible light wavelength $\lambda,$ for each the three types of retinal cones, and let $x(\lambda)$ denote the optical energy of a test stimulus, then the type $i$-cone {\em activation coefficient} is given by
 \begin{equation}\alpha_i(x) = \int_a^b S_i(\lambda)x(\lambda)d\lambda.
 \end{equation}
 \begin{definition} Two color visible lights $x,y$ are said to be {\em metameric} if $\alpha_i(x)=\alpha_i(y), \forall i=1,2,3.$
 \end{definition}
Metamerism is an equivalence relationship $\sim$ on $L^2(\Lambda),$ and the space of perceived light according to Schr$\ddot{o}$dinger(1920)\cite{Sc:1920}
is the space of equivalence classes $\mathcal P = {L^2(\Lambda)/\sim}$ (see Provenzi(2018)\cite{Pr:2018}.
It was assumed that the operations on $L^2(\Lambda)$ lead to certain operators on $\mathcal{P}$, a space thus endowed with an internal associative and commutative operator $\oplus$ having a zero element $0$ and an external multiplication
with positive scalars, $(\alpha, x)\to \alpha \odot x$, that verify the following five axioms:

\begin{axiom} (see Newton(1704)\cite{Ne:1740}): If $x \in \mathcal{P}$ and $\alpha \in \mathbb{R}_+$ , then $\alpha \odot x \in \mathcal{P};$
\end{axiom}

From this axiom, a perceived color has degree of intensity.

\begin{axiom}
If $x \in \mathcal{P}$ then it does not exist any $y \in \mathcal{P}$ such that $x\oplus y = 0;$
\end{axiom}
This axiom means that $\mathcal{P}$ does not have one dimensional subspaces.
\begin{axiom}(see
Grassmann (1853)\cite{Gr:1853} , Helmholtz(1867)\cite{He:1867}) : for every
$ x,y \in \mathcal P $ and for every $\alpha \in [0,1],$ $(\alpha \odot x) \oplus ((1- \alpha )\odot y) \in \mathcal P;$
\end{axiom}
This axiom means that $\mathcal{P}$ is a convex set.

\begin{axiom}(see Helmholtz(1867)\cite{He:1867}): $\forall \{x_k,k=1,2,3,4 \} \subset \mathcal{P},\exists \alpha_k \in \mathbb{R}_+ $ such that $(\alpha_1\odot x_1)\oplus(\alpha_2\odot x_2)\oplus(\alpha_3\odot x_3)\oplus(\alpha_4\odot x_4)=0 .$
\end{axiom}

If $V$ is the vector space spanned by $(\mathcal P, \oplus, \odot),$ the fourth axiom emphasizes that perceived light as a manifold modeled on $V$ has dimension 3 or smaller.

Resnikoff(1074)\cite{Re:1974} added a fifth axiom, concerning the \textbf{local homogeneity} of $\mathcal P$ with respect to changes of background illumination of the visual scene.
 \begin{axiom}
 $\mathcal P$ is locally homogeneous with respect to changes of background.
illumination.
\end{axiom}
We begin we define local homogeneity.
\begin{definition}
A manifold $\mathcal M$ is \textbf{locally homogeneous} with respect to the group $\mathcal{G} $, and a local group action $\alpha$ of $\mathcal{G} $ on $\mathcal M$, that this is locally transitive: for every $x, y \in \mathcal M $ there are open neighborhoods $U_x$ of $x$ and $U_y$ of $y$ , and element $g\in \mathcal G,$ such that $\alpha_g: U_x \to U_y$ given by $\alpha_g(x)=\alpha(g,x)$ is a diffeomorphism.
\end{definition}
If in addition, if $\mathcal M$ has a Riemannian manifold structure, and $\mathcal G$ is a group of local isometries, we say that $\mathcal M$ is a locally homogeneous Riemannian manifold. In dimension 3, and locally homogeneous Riemannian manifold is locally isometric to a homogeneous Riemannian space (for a definition and proof, see Patrangenaru(1996)\cite{Pa:1996}).

The motivation for Resnikoff's axiom is that any perceived light $x \in \mathcal{P}$  can be transformed in a perceived light $y \in \mathcal{P}$ not too different from $ x $ by a change of background illumination, and this process is reversible. We will denote $\alpha_g(x)$ simply by $g(x)$, following Resnikoff's notation. Essentially this kind of locally transformation as a group action, where the group is :
\begin{equation}
GL(P)=\{g \in GL_+(V) | g(x) \in \mathcal P, \forall x \in \mathcal P  \},
\end{equation}
where $V$ is a three dimensional real vector space, to abide by Axiom 4, and $GL_+(V)=\{g\in GL(V), det(g)>0\}$ (see Provenzi(2016)\cite{Pr:2016}.
We assume the perceptual color space is a manifold $\mathcal{M}$ with a metric $\rho$ .
Provenzi(2016, op.cit.) gave his in depth analysis and suggestions for improvement of Resnikoff's 3D homogeneous space models for trichromatic human perception of colors (see Resnikoff(1974)\cite{Re:1974}). Out of the eight 3{D} geometries in Thurston's geometrization conjecture (see Thurston (1982)\cite{Th:1982}), up to an isometry, only two, $E^3$ and $H^2\times E^1,$ satisfy Resnikoff's axioms for the perceived colors space.

Local homogeneity is a very useful property that can be used when conducting statistical analysis on $P,$ in particular for two sample tests. Indeed, if $P$ is a homogeneous space, that admits a simply transitive group $G$, given two means $\mu_1, \mu_2$, there is a unique group element $g = g(\mu_1, \mu_2),$ such that $g \mu_1=\mu_2,$ therefore the null hypothesis $H_0: \mu_1=\mu_2$ is in this case equivalent to the hypothesis $H'_0:g(\mu_1, \mu_2)=1_G,$ where $1_G$ is the identity in the group $G$ (see Osborne et al.(2013)\cite{OPEGS:2013}).

For the remainder of the paper we consider the case the perceptual color space is modeled by $E^3$.
A one to one mapping, $h$, for from the scaled RGB color space,$(0,1)\times (0,1)\times (0,1)$, to $(\mathbb R^+)^3$, (X,Y,Z), could be defined as $log$ transformation, that is

\begin{equation}
h(u,v,w)=-(\ln u,\ln v,\ln w)=(x,y,z),
\end{equation}
where u, v, w are  3 values from the scaled R, G, B axes. Then $(X,Y,Z)$ space satisfied the 4 group axioms and form a group.

We can def group action on  $ (X,Y,Z)$  as $G(\varphi, g)$, where $\varphi$ is element wised multiplication of any point from $(X,Y,Z)$, and $g$ is the the group $(X,Y,Z)$

\begin{equation}
(x',y',z') \cdot (x,y,z) =(x'x,y'y,z'z),
\end{equation}
where $(x',y',z')$ is a point from $g$. The identity element is $(1,1,1)$

Group action $\varphi: G \times (\mathbb R^+)^3 \to (\mathbb R^+)^3 : (g,x) \to \varphi(g,x) \to g \cdot x$
\section{Application to Image Data}
The null hypothesis $H_0: \mu_1=\mu_2$ is in this case equivalent to the hypothesis $H'_0:g(\mu_1, \mu_2)=1_G,$ where $1_G$ is the identity in the group $G$.
$h:((0,1)^3, \ast) \to (\mathbb R^3,+)$ is a group isomorphism. The identity element on $(\mathbb R^3,+)$ is $0_{\mathbb R^3}$. The identity element on $h:((0,1)^3, \ast)$ is $(\frac{1}{e},\frac{1}{e},\frac{1}{e})$.
$H_0: \mu_0=\mu_1$. In $R^3$, we need test if $h(\mu_0)=\mu_1$ or $h(\mu_0,\mu_1)=0_{\mathbb R^3}$.  % $ln(-ln)$ is monotonically decreasing function.

\subsection{Changes in appearance of a scene due to changes in light stimuli.}
\begin{figure}[htbp]
\label{fig:9_times}
\centering
\includegraphics[scale=0.35]{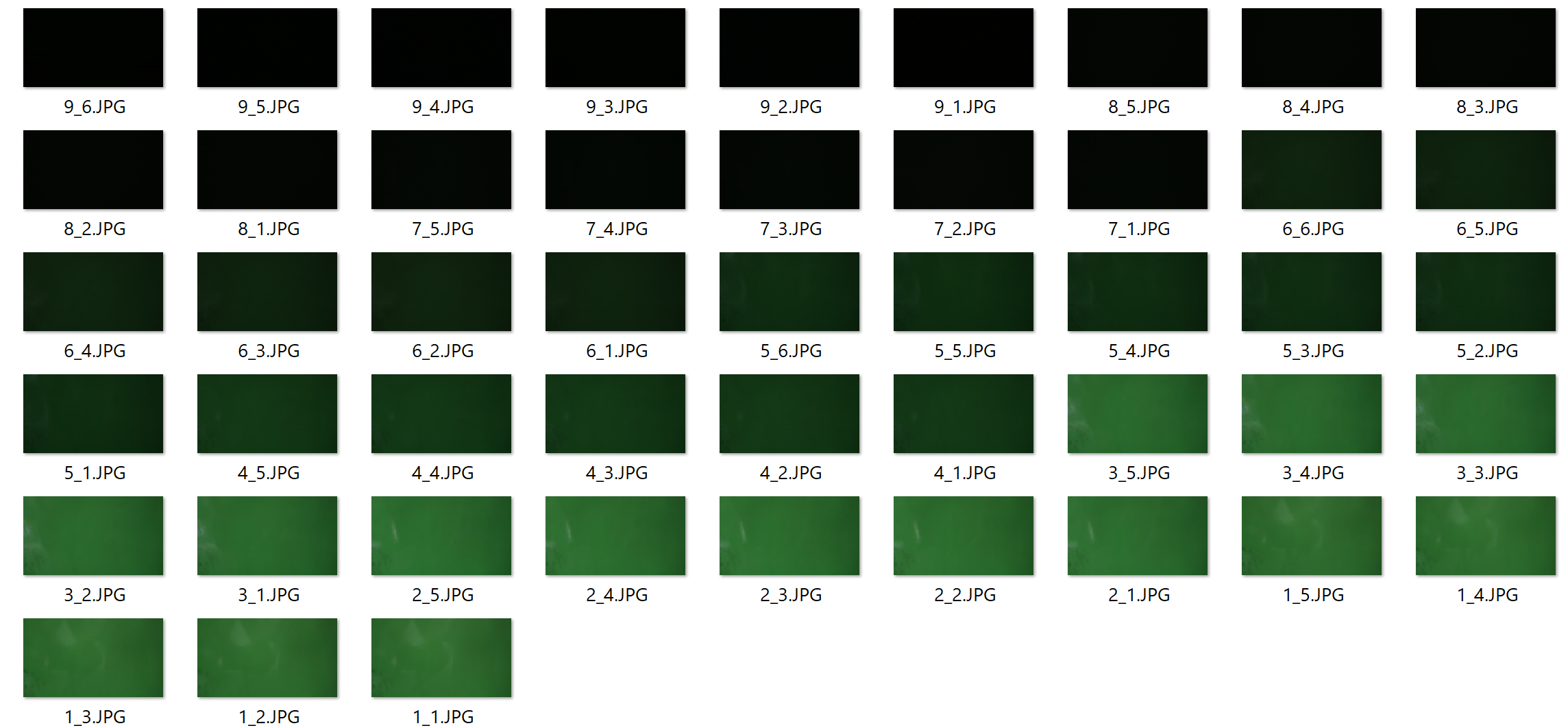}
\caption{Photos of green board taken many times on one day, in increasing temporal order}
\label{Pictures of green board in different times.}
\end{figure}

\begin{figure}[htbp]
\label{fig:var_mean}
\centering
\includegraphics[scale=0.4]{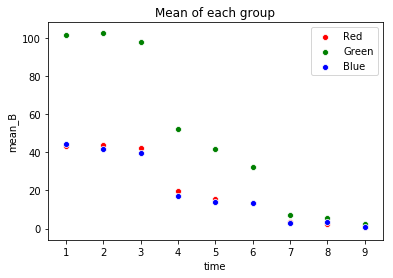}
\includegraphics[scale=0.4]{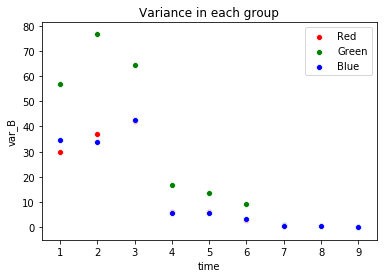}
\caption{Left: Means of R,G,B values of each groups. Variances of R,G,B values within each groups. The R,G,B values in sRGB are scaled from $(0,1)\to(0,255)$.}
\end{figure}

 We considered analyzing differences of perceived daylight, under various conditions, as suggested by McAdams(1942) \cite{Ma:1942}. Collecting perceived light data in human vision is a difficult task. We randomly collected colored digital images instead. Data are collected by taking pictures of a green board from 5pm to 8pm exposed under daylight. We use the mean $[X_r,X_g,X_B]$ to represent each colored image, where $X_r,X_g,X_B$ and the values in red, green, and blue channels. The steps between the time period is 20 minutes. We take five pictures in each time period. There are 9 time period been recorded. Figure \ref{fig:9_times} shows the photo are dimmed from the beginning to the end. Figure \ref{fig:var_mean} show the mean and variance of $[X_r,X_g,X_B]$ within each group.

To measure the statistical significance between each group, we use $h$ transformations for the mean values of each picture and run the Hotelling $T^2$ pairwise tests between each 2 groups images on transformed the data via the $log\circ -log$ map in each fundamental color component, and the, went back to the un-transformed data, multivariate normality for the mean color for each colored image. The results can be found in Table 1 and Table 2.

\begin{table}[htbp]
\label{tb:un_transf}
\small
\begin{tabular}{|c|c|c|c|c|c|c|c|c|c|}
\hline
Group No. & 0            & 1             & 2            & 3             & 4             & 5            & 6             & 7            & 8             \\ \hline
0         & 0.000000     & 30.00     & 185.13 & 2754.86   & 4873.40   & 4752.85  & 9690.22   & 10157.13 & 11579.36  \\ \hline
1         & 30.00    & 0.00      & 444.16 & 36407.01  & 55670.36  & 29836.58 & 181911.71 & 150699.5 & 84816.23  \\ \hline
2         & 185.13   & 444.16    & 0.00 & 91921  & 149665 & 36527 & 870037 & 1110539 & 441494.441 \\ \hline
3         & 2754.86  & 36407.01  & 91921 & 0.00      & 4640.88   & 4127.77  & 103152.53 & 75261.59 & 37563.44  \\ \hline
4         & 4873.40  & 55670.36  & 149665 & 4640.89   & 0.00      & 1187.20  & 83468.18  & 48363.17 & 25371.96  \\ \hline
5         & 4752.85  & 29836.58  & 36527 & 4127.77   & 1187.20   & 0.00     & 5793.04   & 5854.63 & 5741.07   \\ \hline
6         & 9690.22  & 181911.71 & 870037 & 103152.5 & 83468.18  & 5793.04  & 0.00      & 141.50 & 310.38    \\ \hline
7         & 10157.13 & 150699.5 & 1110539 & 75261.59  & 48363.17  & 5854.63  & 141.50    & 0.00 & 923.59    \\ \hline
8         & 11579.36 & 84816.23  & 441494 & 37563.44  & 25371.96  & 5741.07  & 310.38    & 923.59 & 0.00      \\ \hline
\end{tabular}
\caption{Hoteliing $T^2$ tests for pairwise differences RGB values of the green bar scenes}
\end{table}

\begin{table}[htbp]
\label{tb:transf}
\small
\begin{tabular}{|c|c|c|c|c|c|c|c|c|c|}
\hline
Group No. & 0        & 1         & 2          & 3         & 4         & 5        & 6         & 7          & 8         \\ \hline
0         & 0.00     & 30.00     & 185.14     & 2754.87   & 4873.41   & 4752.86  & 9690.22   & 10157   & 11579  \\ \hline
1         & 30.00    & 0.00      & 444.17     & 36407.01  & 55670.37  & 29836.59 & 181911 & 150699  & 84816  \\ \hline
2         & 185.14   & 444.17    & 0.00       & 91920.67  & 149665 & 36527.01 & 870036 & 1110539 & 441494 \\ \hline
3         & 2754.87  & 36407.01  & 91920.67   & 0.00      & 4640.89   & 4127.77  & 103152 & 75261   & 37563.44  \\ \hline
4         & 4873.41  & 55670.37  & 149665  & 4640.89   & 0.00      & 1187.21  & 83468  & 48363.17   & 25371.96  \\ \hline
5         & 4752.86  & 29836.59  & 36527.01   & 4127.77   & 1187.21   & 0.00     & 5793.04   & 5854.63    & 5741.07   \\ \hline
6         & 9690.22  & 181911 & 870036  & 103152 & 83468.18  & 5793.04  & 0.00      & 141.50     & 310.38    \\ \hline
7         & 10157.13 & 150699 & 1110539 & 75261.59  & 48363.17  & 5854.63  & 141.50    & 0.00       & 923.59    \\ \hline
8         & 11579.36 & 84816.23  & 441494  & 37563.44  & 25371.96  & 5741.07  & 310.38    & 923.59     & 0.00      \\ \hline
\end{tabular}
\caption{Hoteliing $T^2$ tests for pairwise differences RGB values of the green bar scene after the h transformation }
\end{table}
\normalsize
\subsection{Detection of differences between indoor and outdoor light conditions R, G and B perceived images }
\begin{figure}[htbp]
\centering
\label{fig:3boards}

\includegraphics[scale=0.7]{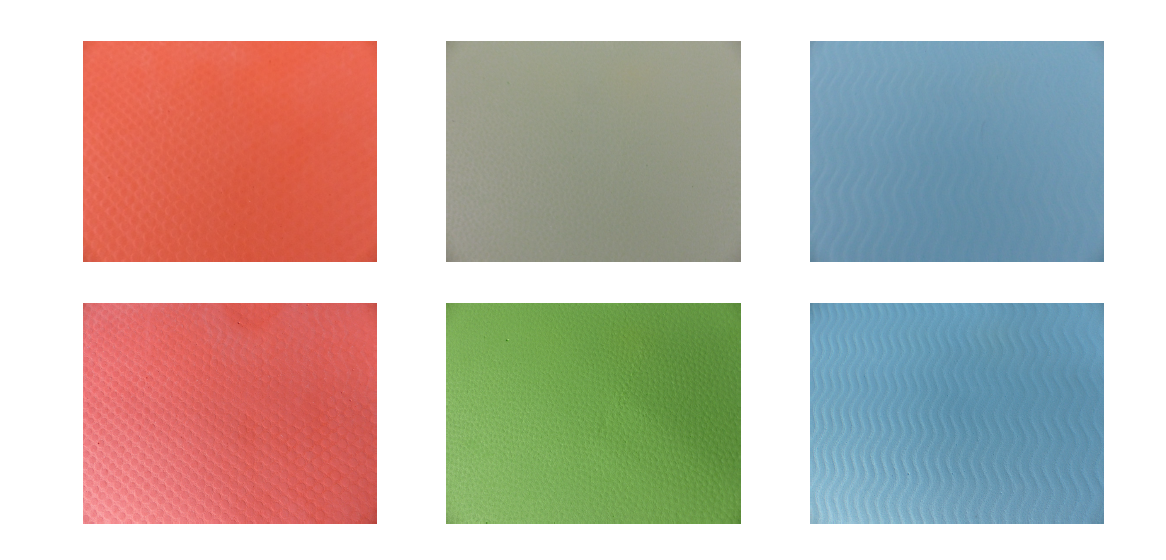}

\caption{Photos in first row are taken indoors. The second row pictures are taken outdoors.}
\end{figure}

\begin{figure}[htbp]
\centering
\label{fig:sct_3boards}
\includegraphics[scale=0.5]{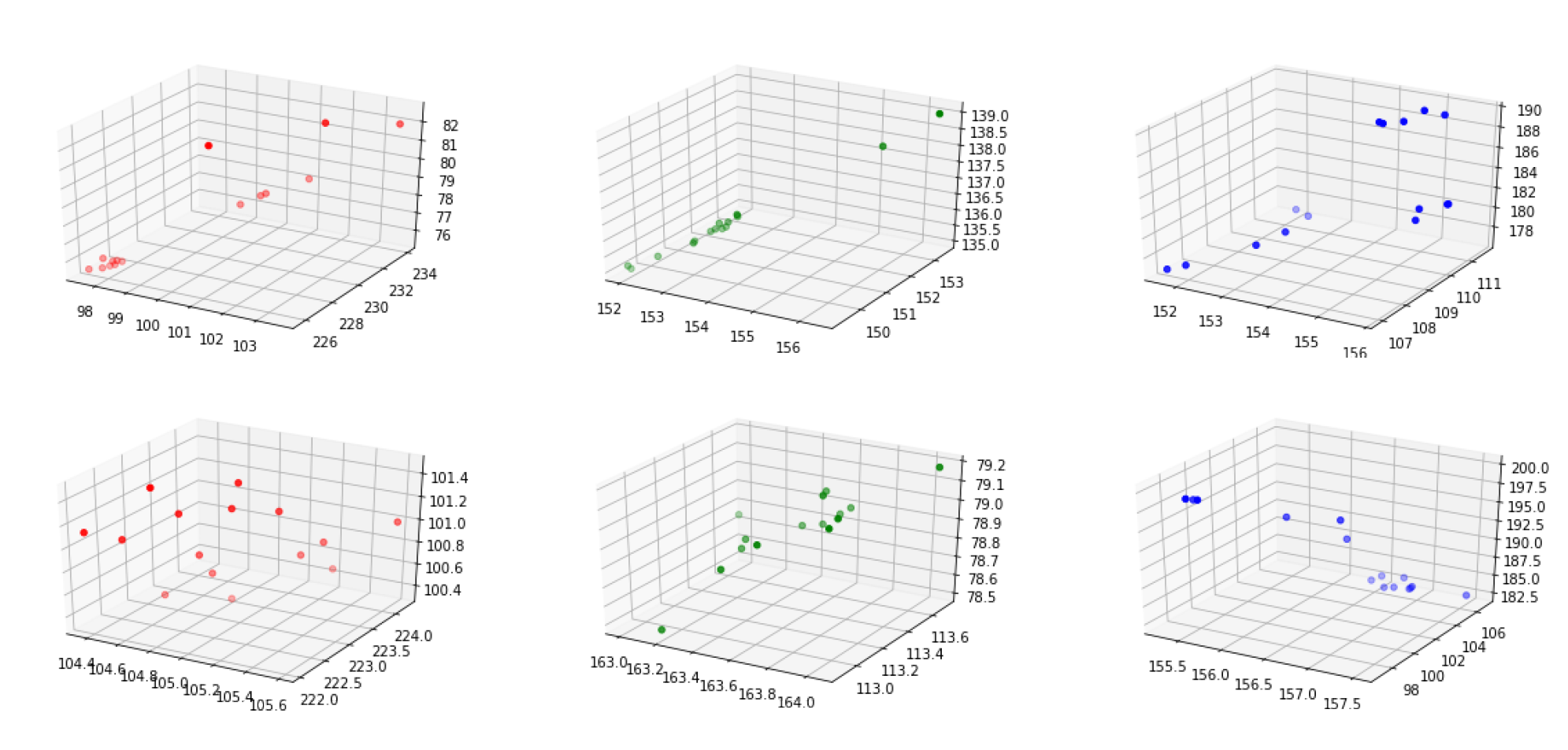}
\caption{Upper row-3D representation of the images taken indoors. Lower row-3D representation of the images taken outdoors.}
\end{figure}

The object we are interested in color detection for colored scenes that are perceived as red, green, and blue respectively. First we exposed these scenes under natural daylight and took fifteen photos for each of the scenes. Then we changed the light condition to indoor, and repeated the image collection under these light conditions. We have six groups of data. Figure \ref{fig:3boards} shows how is the sample looks like in each group. The difference between different color board can be easily distinguished by the human eye. The differences between indoor and outdoor are not that clear though (the difference between indoor and outdoor blue board images are hard to distinguish for example). The representation of each in 3D vectorized mean image $[X_r,X_g,X_B]$, where each element is the value in relevant color channel is displayed in a scatter plot of the mean representation for the six groups can be found in figure \ref{fig:sct_3boards}.

To see the statistical significance between groups, we transformed($log\circ -log$) the data first then calculate the Hotelling $T^2$ tests, assuming multivariate normality for the 3D data (given that these are averages of pixel RGB over the entire board). The result is given in table \ref{tb:3boards}
\begin{table}[]
\caption {Values of $T^2$ statistic for indoor and outdoor color board}
\label{tb:3boards}
\centering
\begin{tabular}{|l|l|l|l|}
\hline
Indoor\textbackslash{}Outdoor & Red          & Green        & Blue          \\ \hline
Red                           & 1.773347e+04 & 1.178999e+06 & 170012.841702 \\ \hline
Green                         & 2.977303e+06 & 8.608758e+05 & 5349.027276   \\ \hline
Blue                          & 6.861927e+04 & 5.631528e+03 & 68.098009     \\ \hline
\end{tabular}
\end{table}
\section{Discussion and Future Work}
The role of color perception in human and machine vision can not be overstated. In this paper we have used only one of the two
homogeneous space models for the space of perceived colors. Future work will be dedicated to an analysis on the non-Euclidean Lie group
structure of this object space. Note that RGB based machine vision, emulating human vision was successfully used in RGB {3D} surface reconstruction from color digital camera images, simplifying 3D landmark registration and projective analysis of 3D scenes (see Patrangenaru et al.(2016)\cite{PaYaBa:2016}).


\begin{thebibliography}{22}
%
\bibitem{Fa:2005} M.D. Fairchild. (2005). {\em Color appearance models}. {Wiley}.
%
\bibitem{Gr:1853} H.G. Grassmann (1853). Zur Theorie der Farbmischung. {\em Poggendork's Annalen der
Physik,}{\bf 89}, 69--84.
%
\bibitem{Gu:1932} J. Guild(1932). The colorimetric properties of the spectrum. {\em Philosophical Transactions of the Royal Society of London. Series A, Containing Papers of a Mathematical or Physical Character}. {\bf 230},149–-187.
%
\bibitem{He:1867} H. von Helmholtz (1867). {\em Handbuch der physiologischen Optik}, {Allgemeine Ency-
clopdie der Physik, IX. Band. Leipzig, Leopold Voss}.
%
\bibitem{Re:1974} H. L. Resnikoff,
{Differential geometry and color perception}, {\em J. Math. Biol.} {\bf 1}, no. 2, (1974/75), 97--131
%
\bibitem{MaDa:2008} Margaret Livingstone, David H Hubel (2008). {\em Vision and art : the biology of seeing}.  New York : Abrams.
%
\bibitem{Ma:1942} D.L. MacAdams (1942). Visual sensitivities to colour differences in daylight, {\em Journal of the Optical Society of America A,} {\bf 32} (5), 247--274.
%
\bibitem{Ne:1993} Isaac Newton(1993). A new theory about light and colors. {\em Amer. J. Phys.} {\bf 61} , no. 2, 108–-112.
%
\bibitem{Ne:1740} I. Newton (1704). {\em Opticks}, London, Smith and Walford.
%
\bibitem{Pa:1996} V. Patrangenaru (1996). Classifying 3- and 4-dimensional homogeneous Riemannian manifolds by Cartan triples. {\em Pacific J. Math.} {\bf 173},  511-–532.
%
\bibitem{OPEGS:2013} D. Osborne, V. Patrangenaru, L. Ellingson, D. Groisser
and A. Schwartzman. (2013). Nonparametric Two-Sample Tests on Homogeneous Riemannian Manifolds, Cholesky Decompositions and Diffusion Tensor Image Analysis. {\em  Journal of Multivariate Analysis}.  {\bf 119}, 163-175.

\bibitem{PaYaBa:2016} Patrangenaru, V.; Yao, K. D.; Balan, V.(2016).
{3D} face analysis from digital camera images. {\em  Proceedings—The International Conference of Differential Geometry and Dynamical Systems (DGDS-2015), BSG Proc.}, {\bf 23,} 43–-55.
%
\bibitem{Pr:2018} Edoardo Provenzi(2018). {\em Color Image Processing.}https://www.math.u-bordeaux.fr/$\sim$eprovenzi/include/Notes$\_$main.pdf
%
\bibitem{Pr:2016} Edoardo Provenzi(2016).
A differential geometry model for the perceived colors space. {\em
Int. J. Geom. Methods Mod. Phys.} {\bf 13}, no. 8, 1630008, 8 pp.
%

\bibitem{Sc:1920}E. Schr$\ddot{o}$dinger (1920). {Grundlinien einer Theorie der Farbenmetrik im Tagessehen
(Outline of a theory of colour measurement for daylight vision)}, {\em Annalen
der Physik} {\bf 63 }(4), 397-456; 481-520. Available in English in Sources
of Colour Science, Ed. David L. MacAdam, The MIT Press (1970), 13482.
%
\bibitem{Th:1982} William P. Thurston(1982).
Three-dimensional manifolds, Kleinian groups and hyperbolic geometry.
{\em Bull. Amer. Math. Soc. (N.S.)} {\bf 6}, no. 3, 357–-381.
%
\bibitem{We:1999} IEC Webstore(1999). Multimedia systems and equipment-Colour measurement and management-Part 2-1: Colour management-Default RGB colour space-sRGB.
%
\bibitem{Wr:1928}William David Wright(1928). A re-determination of the trichromatic coefficients of the spectral colors.{\em Transactions of the Optical Society.} {\bf 30 (4)}, 141–-164.
%



\end{thebibliography}
\end{document}